\documentclass{article}

\usepackage{microtype}
\usepackage{graphicx}
\usepackage{subfigure}
\usepackage{booktabs} % for professional tables

\usepackage{hyperref}
\usepackage{wrapfig}
\usepackage{subcaption}
\usepackage{booktabs}
\usepackage{caption}
\usepackage{enumitem}
\usepackage{url}
\usepackage{bbm}
\usepackage{multirow}

\usepackage[accepted]{icml2024}

\usepackage{amsmath}
\usepackage{amssymb}
\usepackage{mathtools}
\usepackage{amsthm}
\usepackage{bm} 
\usepackage[capitalize,noabbrev]{cleveref}
\theoremstyle{plain}

\theoremstyle{definition}

\theoremstyle{remark}

\usepackage[textsize=tiny]{todonotes}
\usepackage{tabularx}

\icmltitlerunning{Improved Operator Learning by Orthogonal Attention}

\begin{document}

\twocolumn[
\icmltitle{Improved Operator Learning by Orthogonal Attention}



\icmlsetsymbol{equal}{*}

\begin{icmlauthorlist}
\icmlauthor{Zipeng Xiao}{sjtu}
\icmlauthor{Zhongkai Hao}{thu}
\icmlauthor{Bokai Lin}{sjtu}
\icmlauthor{Zhijie Deng}{sjtu}
\icmlauthor{Hang Su}{thu}
\end{icmlauthorlist}

\icmlaffiliation{sjtu}{Qing Yuan Research Institute, SEIEE, Shanghai Jiao Tong University}
\icmlaffiliation{thu}{Dept. of Comp. Sci. \& Tech., Tsinghua University}
\icmlcorrespondingauthor{Zhijie Deng}{zhijied@sjtu.edu.cn}
\icmlcorrespondingauthor{Hang Su}{suhangss@tsinghua.edu.cn}

\icmlkeywords{Neural Operator, Attention, Transformer}

\vskip 0.3in
]



\printAffiliationsAndNotice{} % otherwise use the standard text.

\begin{abstract}
% Neural operators, as an efficient surrogate model for learning the solutions of PDEs, have received extensive attention in the field of scientific machine learning.
% Among them, attention-based neural operators have become one of the mainstreams in related research. 
% Despite the considerable expressiveness, existing approaches may face generalization challenges, especially when faced with limited training data. %\hangx{strong expressive power/capability is some positive words, and we could use some neutral or negative words to express the limitation}. 
This work presents orthogonal attention for constructing neural operators to serve as surrogates to model the solutions of a family of Partial Differential Equations (PDEs). 
The motivation is that the kernel integral operator, which is usually at the core of neural operators, can be reformulated with orthonormal eigenfunctions. 
Inspired by the success of the neural approximation of eigenfunctions~\cite{deng2022neuralef}, we opt to directly parameterize the involved eigenfunctions with flexible neural networks (NNs), based on which the input function is then transformed by the rule of kernel integral. 
Surprisingly, the resulting NN module bears a striking resemblance to regular attention mechanisms, albeit without softmax. 
Instead, it incorporates an orthogonalization operation that provides regularization during model training and helps mitigate overfitting, particularly in scenarios with limited data availability. 
In practice, the orthogonalization operation can be implemented with minimal additional overheads.
Experiments on six standard neural operator benchmark datasets comprising both regular and irregular geometries show that our method can outperform competing baselines with decent margins.
\end{abstract}

\section{Introduction}
Partial Differential Equations (PDEs) are essential tools for modeling and describing intricate dynamics in scientific and engineering domains~\cite{zachmanoglou1986introduction}. 
Solving the PDEs routinely rely on well-established numerical approaches such as finite element methods (FEM)~\cite{zienkiewicz2005finite}, finite difference methods (FDM)~\cite{thomas2013numerical}, spectral methods~\cite{ciarlet2002finite,courant1967partial}, etc. % are widely employed. These numerical solvers are extensively evaluated and applicable to a wide range of PDE problems.
Due to the infinite-dimensional nature of the function space, traditional numerical solvers often rely on discretizing the data domain. However, this introduces a balance between efficiency and accuracy: finer discretization offers higher precision but at the expense of greater computational complexity.
% Since the function space is infinite-dimensional, these classic numerical solvers typically proceed by discretizing the data domain, which, however, introduces a trade-off between efficiency and efficacy---achieving high precision necessitates a fine discretization and hence leads to increased computational complexity. 
% Their accuracy is intricately tied to the discretization of the data domain, where fine discretization routinely means high precision. 
% As a result, such solvers may manifest sluggishness and inefficiency when tasked with attaining precise outcomes.

Deep learning methods have shown promise in lifting such a trade-off \cite{li2020fourier} thanks to their high inference speed and expressiveness. 
%efficacy in diverse tasks and are anticipated to provide intermediate approximations to Partial Differential Equations (PDEs) solutions. 
Specifically, physics-informed neural networks (PINNs)~\cite{raissi2019physics} first combine neural networks (NNs) with physical principles for PDE solving. 
Yet, PINNs approximate the solution associated with a certain PDE instance and hence cannot readily adapt to problems with different yet similar setups. 
By learning a map between the input condition and the PDE solution in a data-driven manner, neural operators manage to solve a family of PDEs, with the DeepONet \cite{lu2019deeponet} as a representative example. %, which have prompted the emergence of a broad range of research works. 
% DeepONet \cite{lu2019deeponet} proposes a model to learn the solution operator and attributes the aptitude to the universal approximation theorm. 
Fourier Neural Operators (FNOs)~\cite{li2020fourier,tran2023factorized,li2022fourier,wen2022u,grady2022towards,gupta2021multiwavelet,xiong2023koopman} shift the learning to Fourier space to enhance speed while maintaining efficacy through the utilization of the Fast Fourier Transform (FFT). %\zhijie{to achieve what?}. 
Since the development of attention mechanism~\cite{vaswani2017attention}, considerable effort has been devoted to developing attention-based neural operators to improve expressiveness and address irregular mesh~\cite{cao2021choose,li2023transformer,ovadia2023vito,fonseca2023continuous,hao2023gnot,li2023scalable}.

Despite the considerable progress made in neural operators, there remain non-trivial challenges in its practical applications. 
On the one hand, the training targets of neural operators are usually acquired from classical PDE solvers, which can be computationally demanding. 
For instance, simulations for tasks like airfoils can require about 1 CPU-hour per sample~\cite{li2022fourier}. 
On the other hand, complex deep models are prone to deteriorate when confronted with limited training data.

This work aims to develop a novel neural operator that inherently accommodates proper regularization to cope with the challenges in the processing of PDE data. 
We start from the observations that the kernel integral operator, a core module of the solving operator of PDEs, can be rewritten with orthonormal eigenfunctions.
% \zhijie{say that the kernel integral operator in Green function method, which works for some simplified pdes, can be represented by orthonormal eigenfunctions by Mercer's theorem. }
% originates from the kernel integral operator and Green function. 
% We formulate the kernel by leveraging a multitude of inherent orthonormal eigenfunctions associated with the integral operators. 
Such an expansion substantially resembles the attention mechanism without softmax while incorporating the orthogonal regularization (detailed in Section~\ref{sec:theo}). 
% This integral operator possess the capacity to effectively approximate the solution operators of PDEs and exhibit resemblances to attention structure without softmax and normalization. 
Empowered by this, we follow the notion of neural eigenfunctions~\cite{deng2022neuralef,deng2022neural} to implement an orthogonal attention module and stack it repeatedly to construct \textit{orthogonal neural operator (ONO)}. 
% Drawing from this perspective, we present a novel attention mechanism characterized by linear complexity and orthogonalization applied to the learned basis functions. 
% Through this approach, the model gains enhanced capacity to approximate the kernel while mitigating overfitting. This is attributed to the normalization-like effect and the regularization induced by orthogonalization. 
% Expanding on this method, we subsequently propose a flexible and scalable framework for the solution operators of PDEs, \textbf{Orthogonal Neural Operator (ONO)}. 
% ONO employs a hierarchical network architecture, thereby bolstering expressiveness while avoiding the complexity associated with heavy models.
As shown in Figure~\ref{fig:ono}, ONO is structured with two disentangled pathways. 
The bottom one approximates the eigenfunctions through expressive NNs, while the top one specifies the evolvement of the PDE solution. 
In practice, the orthogonalization operation can be implemented by cheap manipulation of the exponential moving average (EMA) of the feature covariance matrix.
It is empirically proven that ONO can generalize substantially better than competitive baselines across both spatial and temporal axes. %, reducing the prediction error by up to $80 \%$ on the Darcy benchmark. 
To summarize, our contributions are:
\begin{itemize}
    \vspace{-5pt}
    \setlength{\itemsep}{5pt}
    \item We introduce the novel orthogonal attention, which is inherently integrated with orthogonal regularization while maintaining moderate complexity, and detail the theoretical insights. %, supported by theorem-based interpretations. This operator functions as an kernel integral operator and offers effective regularization for the model.
    \item We introduce ONO, a neural operator built upon orthogonal attention. ONO employs two disentangled pathways for approximating the eigenfunctions and PDE solutions separately. 
    \item We conduct comprehensive experiments on six challenging operator learning benchmarks and achieve satisfactory results: ONO reduces prediction errors by up to $30 \%$ compared to baselines and achieves $80 \%$ reduction of test error for zero-shot super-resolution on Darcy.
\end{itemize}

\section{Related Work}
\subsection{Neural Operators}
%Neural operators eliminate the need for retraining by learning PDE solutions across various discretizations, without requiring prior knowledge of the underlying PDE.
Neural operators map infinite-dimensional input and solution function spaces, allowing them to handle multiple PDE instances without retraining.
Following the advent of DeepONet~\cite{lu2019deeponet}, the domain of operator learning
% learning employing neural networks 
has recently gained much attention.
Specifically, DeepONet employs a branch network and a trunk network to separately encode input functions and location variables, subsequently merging them for output computation. 
%Physics-informed DeepONet~\cite{wang2021learning} introduces an extension of DeepONet that incorporates prior knowledge of PDEs and demonstrates the capacity for training without any training data.
Numerous alternative variants have been proposed from various perspectives thus far ~\cite{grady2022towards,wen2022u,xiong2023koopman}.
FNO~\cite{li2020fourier} learns the integral operator in the spectral domain to conjoin accuracy and inference speed. 
Geo-FNO~\cite{li2022fourier} employs a map connecting irregular domains and uniform latent meshes to address arbitrary geometries effectively. 
% Recently, numerous FNO variants have been introduced.
F-FNO~\cite{tran2023factorized} improves FNO by integrating dimension-separable Fourier layers and residual connections.
However, FNOs are grid-based, leading to increased computational demands for both training and inference as PDE dimensions expand. 

Considering the input sequence as a function evaluation within a specific domain, attention operators can be seen as learnable projection or kernel integral operators.
These operators have gained substantial research attention due to their scalability and effectiveness in addressing PDEs on irregular meshes. 
\citet{kovachki2021neural} demonstrates that the standard attention mechanism can be considered as a neural operator layer.
Galerkin Transformer~\cite{cao2021choose} proposes two self-attention operators without softmax and provides theoretical interpretations for them.
HT-Net~\cite{liu2022ht} proposes a hierarchical attention operator to solve multi-scale PDEs.
%OFormer~\cite{li2022transformer} proposes a cross-attention module enabling distinct discretization between input and output functions.
GNOT~\cite{hao2023gnot} proposes a linear cross-attention block to facilitate the encoding of diverse input types.
However, despite their promising potential, attention operators are susceptible to overfitting, especially when the available training data are rare.

\subsection{Efficient Attention Mechanisms}
The Transformer model~\cite{vaswani2017attention} has gained popularity in diverse domains~\cite{chen2018best,parmar2018image,rives2021biological}.
However, the vanilla softmax attention encounters scalability issues due to its quadratic space and time complexity. 
To tackle this, several methods with reduced complexity have been proposed~\cite{child2019generating,zaheer2020big,wang2020linformer,katharopoulos2020transformers,xiong2021nystromformer}. 
Concretely, Sparse Transformer~\cite{child2019generating} reduces complexity by sparsifying the attention matrix.
Linear Transformer~\cite{katharopoulos2020transformers} achieves complexity by replacing softmax with a kernel function.
Nystr{\"o}mformer~\cite{xiong2021nystromformer} employs the Nystr{\"o}m method to approximate standard attention, maintaining linear complexity.

In the context of PDE solving, 
%attention mechanisms without softmax are introduced, featuring reduced computational complexity.
Galerkin Transformer~\cite{cao2021choose} proposes the linear Galerkin-type attention mechanism, which can be regarded as a trainable Petrov–Galerkin-type projection.
OFormer~\cite{li2023transformer} develops a linear cross-attention module for disentangling the output and input domains.
FactFormer~\cite{li2023scalable} employs axial computation in the attention operator to reduce computational costs.
Compared to them, we not only introduce an attention mechanism without softmax at linear complexity but also include an inherent regularization mechanism.

\section{Methodology}
This section begins with an overview of the orthogonal neural operator and subsequently delves into the orthogonal attention mechanism and its theoretical foundations.

\label{sec:mode}
\begin{figure*}[t]
    \begin{center}
    %\framebox[4.0in]{\;}
    \includegraphics[width=0.83\textwidth]{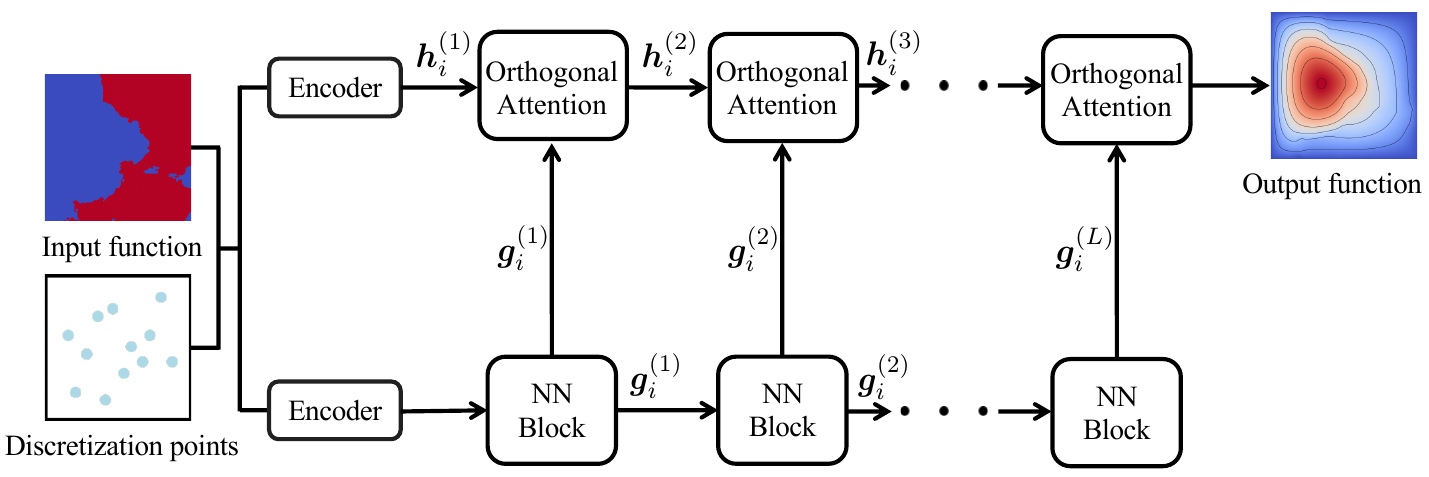}
    \end{center}
    %\captionsetup{justification=centering}
    \caption{Model overview. There are two flows in ONO. %input function and discretization points are fed into ONO. 
    The bottom one extracts expressive features for input data, forming an approximation to the eigenfunctions associated with the kernel integral operators for defining ONO.
    The top one updates the PDE solutions based on orthogonal attention, which involves linear attention and orthogonal regularization.}%\hangx{could be redrawn to make it more expressive} }
    \label{fig:ono}    
\end{figure*}

\subsection{Problem Setup}
Operator learning involves learning the mapping from the space of input functions $f: D\to \mathbb{R} ^{d_{f}} \in \mathcal{F}$ to the space of PDE solutions $u: D\to \mathbb{R} ^{d_{u}} \in \mathcal{U}$, % as two Banach spaces of the input functions and the solution functions, 
where $D$ is a bounded open set. 
Let $\mathcal{G}:\mathcal{F}\rightarrow\mathcal{U}$ denotes the ground-truth solution operator.
Our objective is to train a $\bm{\theta}$-parameterized neural operator $\mathcal{G}_{\bm{\theta}}$ to approximate $\mathcal{G}$.  
The training is driven by a collection of function pairs $\{f_{i},u_{i}\}_{i=1}^{N}$. %, where fif_{i} follows a distribution μ\mu supported on F\mathcal{F}. 
Deep models routinely cannot accept an infinite-dimensional function as input or output, so we discretize $f_{i}$ and $u_{i}$ on mesh $\mathbf{X} := \{ \bm{x}_{j} \in D \}_{1\leq j \leq M }$, yielding $\bm{f}_i :=\{(\bm{x}_j, f_{i}(\bm{x}_j))\}_{1 \leq j \leq M}$ and $\bm{u}_i := \{(\bm{x}_j, u_{i}(\bm{x}_j) )\}_{1 \leq j \leq M}$.
We use $\bm{f}_{i,j}$ to denote the element in $\bm{f}_i$ that corresponds to $\bm{x}_j$. 
The data fitting is usually achieved by optimizing the following problem:
% Our objective is to identify a solution that minimizes the ensuing optimization problem:
\begin{equation}
\min\limits_{\bm{\theta}}\frac{1}{N}\sum_{i=1}^{N}\frac{\Vert \mathcal{G}_{\bm{\theta}}(\bm{f}_{i})-\bm{u}_{i}\Vert_{2}}{\Vert \bm{u}_{i}\Vert_{2}},
\label{eq:optim}
\end{equation}
where the regular mean-squared error (MSE) is augmented with a normalizer $\Vert \bm{u}_{i}\Vert_{2}$ to account for variations in absolute scale across benchmarks.
We refer to this error as $l_{2}$ relative error in the subsequent sections.

\subsection{Orthogonal Neural Operator }
%\hzk{in the overview, we need to re-state the challenges and our effort to solve the challenge}
\textbf{Overview.} %\zhijie{why sometimes you capitalize the second word, sometimes not?}
Basically, an $L$-stage ONO takes the form of 
\begin{equation}
\label{eq:archi}
    \mathcal{G}_{\bm{\theta}} := \mathcal{P} \circ \mathcal{K}^{(L)} \circ \sigma \circ \mathcal{K}^{(L-1)} \circ \dots \circ \sigma\circ \mathcal{K}^{(1)} \circ \mathcal{E},
\end{equation}
where $\mathcal{E}$ maps $\bm{f}_i$ to hidden states $\bm{h}^{(1)}_i \in \mathbb{R}^{M\times d}$, $\mathcal{P}$ projects the states to solutions, and $\sigma$ denotes the non-linear transformation. 
% between the hidden representations of NNs and the inputs and outputs. 
$\mathcal{K}^{(l)}$ % : \{ h_{l} : D \rightarrow \mathbb{R}^{d_{h_{l}}} \} \mapsto \{ h_{l+1} : D \rightarrow \mathbb{R}^{d_{h_{l+1}}} \}$ 
refer to parameterized kernel integral operators following the prior arts in neural operator~\cite {kovachki2021neural}, which is motivated by the link between kernel integral operator and Green's function for solving linear PDEs. 

Note that $\mathcal{K}^{(l)}$ accepts hidden states $\bm{h}^{(l)}_i \in \mathbb{R}^{M \times d}$ as input instead of infinite-dimensional functions as in the traditional kernel integral operator. 
It should also rely on some parameterized configuration of a kernel. 
FNO addresses this by employing linear transformations on truncated frequency modes in the Fourier domain, albeit with potential limitations in effectively handling high-frequency information~\cite{li2020fourier}.
%\zhijie{repeat some issues of FNO}. 
Instead, we advocate directly parameterizing the kernel in the original space with the help of neural eigenfunctions~\cite{deng2022neuralef,deng2022neural}.
Specifically, we leverage an additional NN to extract hierarchical features from $\bm{f}_i$, which, after orthogonalization and normalization, suffice to define $\mathcal{K}^{(l)}$. 
The orthogonalization serves as a regularization to enhance the model generalization ability.

We outline the overview of ONO in Figure~\ref{fig:ono}, where the two-flow structure is clearly displayed. 
We pack the orthonormalization step and eigenfunctions-based kernel integral into a module named \emph{orthogonal attention}. 
The decoupled architecture offers significant flexibility in specifying the NN blocks within the bottom flow.

\textbf{Encoder.}
The encoder is multi-layer perceptrons (MLPs) that accept $\bm{f}_i$ as input for dimension lifting.
% To combine the information in , t transforms their concatenation with . 
% We use dedicated encoders for the two flows in ONO. 
Features at every position $\bm{x}_j$ are extracted separately. 

% in ONO, which are implemented as . 

% \zhijie{move this sentence to experiment settings}We employ two separate encoders as the lifting L\mathcal{L} to map f(\bm{x})f(\bm{x}) to its hidden representation.
% In practice, these encoders are implemented using individual point-wise multi-layer perceptrons (MLPs).
% The outputs of these two distinct encodings are subsequently fed into the orthogonal attention layer and the NN block.

\textbf{NN Block.}
In the bottom flow, the NN blocks are responsible for extracting features, which subsequently specify the kernel integral operators in the orthogonal attention modules. 
% and forwarding them to the attention layer and the subsequent NN Block.
We can leverage any existing architecture here but focus on transformers due to their great expressiveness.
In detail, we formulate the NN block as follow:
\begin{equation}
\begin{aligned}
    \tilde{\bm{g}}^{(l)}_i &= \bm{g}^{(l)}_i + \mathrm{Attn}(\mathrm{LN}(\bm{g}^{(l)}_i),\\
    \bm{g}^{(l+1)}_i &= \tilde{\bm{g}}^{(l)}_i + \mathrm{FFN}(\mathrm{LN}(\tilde{\bm{g}}^{(l)}_i)),    
\end{aligned}
\end{equation}
where $\bm{g}^{(l)}_i \in \mathbb{R}^{M \times d'}$ denotes the output of $l$-th NN block for the data $\bm{f}_i$. 
$\mathrm{Attn}(\cdot)$ represents a self-attention module applied over the $M$ positions. 
$\mathrm{LN}(\cdot)$ indicates layer normalization~\cite{ba2016layer}. %, a technique commonly used in transformer architectures~\cite{vaswani2017attention}.
$\mathrm{FFN}(\cdot)$ refers to a two-layer feed forward network. 
%\zhijie{put this into the experiment setting: In detail, we choose Linear attention from~\cite{katharopoulos2020transformers} or Nystrom attention from~\cite{xiong2021nystromformer} as the attention mechanism empirically.
%Both of these mechanisms exhibit linear time and space complexity while maintaining remarkable performance.}
Here, we can freely choose well-studied self-attention mechanisms, e.g., standard attention~\cite{vaswani2017attention} and other variants that enjoy higher efficiency to suit specific requirements.%\hzk{here we say our framework is flexible and can freely choose attention, then why we need orthogonal attention? some connecting words are needed} % compatibility with a variety of neural network architectures for the NN Block, such as the standard transformer block~\cite{vaswani2017attention} and other linear complexity variants~\cite{wang2020linformer,cao2021choose}.
% This flexibility allows the selection of the architecture that suits the specific requirements.
\begin{figure}[t]
    \begin{center}
    %\framebox[4.0in]{\;\;}
    \includegraphics[width=\linewidth]{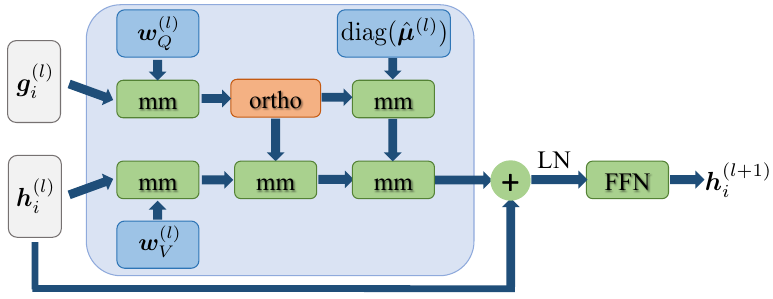}
    \end{center}
    \vspace{-2ex}
    \caption{Orthogonal attention: the module incorporates matrix multiplications (``mm") and an orthogonalization process (``ortho"). The output of the NN block, denoted as $\bm{g}_{i}^{(l)}$, and the hidden state of the input function, represented as $\bm{h}_{i}^{(l)}$, undergo processing as shown in Equation~\ref{eq:attnwithmu}. Following this, the module includes a residual connection, layer normalization, and a two-layer FFN.}
    \label{fig:AttnLayer}    
\end{figure}

\textbf{Orthogonal Attention.} 
We introduce the orthogonal attention module with orthogonal regularization to remediate the potential overfitting in the context of operator learning. 
This module characterizes the evolution of PDE solutions. % and enhances generalization performance.
% In ONO's upper flow, a stack of orthogonal attention layers, as shown in Figure~???????????????\ref{fig:AttnLayer}, map the input functions into the target functions.
It transforms the deep features from the NN blocks to orthogonal eigenmaps, based on which the kernel integral operators are constructed and the hidden states of PDE solutions are updated. 
Concretely, we first project the NN features $\bm{g}^{(l)}_i \in \mathbb{R}^{M \times d'}$ to: % the following kk-dimensional eigenmaps:
% orthogonal attention is computed through:
\begin{equation}
    \hat{\bm{\psi}}^{(l)}_i = \mathrm{ort}(\hat{\bm{g}}^{(l)}_i) = \mathrm{ort}(\bm{g}^{(l)}_i \bm{w}^{(l)}_Q) \in \mathbb{R}^{M \times k},
    %\bm{z} = \mathrm{Attn}_{ort} := \psi(\bm{x})(\psi(\bm{x})^{\top}\bm{y})
\end{equation}
where $\bm{w}^{(l)}_Q \in \mathbb{R}^{d' \times k}$ is a trainable weight. 
$\mathrm{ort}(\cdot)$ is the orthonormalization operation which renders each column of $\hat{\bm{\psi}}^{(l)}_i$ correspond to the evaluation of a specific neural eigenfunction on $\bm{f}_i$. 

Given these, the orthogonal attention update the hidden states $\bm{h}^{(l)}_i$ of PDE solutions via:
\begin{equation}
    \label{eq:attnwithmu}
    \tilde{\bm{h}}^{(l+1)}_{i}= \hat{\bm{\psi}}^{(l)}_i\mathrm{diag}(\hat{\boldsymbol{\mu}}^{(l)})[{\hat{\bm{\psi}}^{(l)}_i}{}^{\top} (\bm{h}^{(l)}_i \bm{w}^{(l)}_V)],
\end{equation}
% Here ort(⋅)\mathrm{ort}(\cdot) an orthogonalization process applied to the features to obtain orthonormal eigenfunctions and 
where $\bm{w}^{(l)}_V \in \mathbb{R}^{d\times d}$ is a trainable linear weight to refine the hidden states, and $\hat{\boldsymbol{\bm{\mu}}}^{(l)} \in \mathbb{R}_+^{k}$ denote positive eigenvalues associated with the induced kernel and are trainable in practice. 
This update rule is closely related to Mercer's theorem, as will be detailed in Section~\ref{sec:theo}. 
%\zhijie{kk is an important hyper-parameter for our method. you should explain its meaning and impact.}
% gl(\bm{x})g_{l}(\bm{x}) and hl−1h_{l-1} signify eigenfunctions from the corresponding NN block and the function representation. 
% For gl(\bm{x})g_{l}(\bm{x}) can
% g(\bm{x})∈Rn×dg(\bm{x}) \in \mathbb{R}^{n \times d} and \bm{y}∈Rn×d\bm{y} \in \mathbb{R}^{n \times d} denote the data features and hidden representation yly_{l} evaluated at the grid points, 
% It then proceeds by:
% \begin{equation}
%     \label{eq:AttnMM}
%     \mathrm{Attn}_{o}(?) := \mathrm{ort}(\mQ)\mathrm{diag}(\hat{\boldsymbol{\mu}})(\mathrm{ort}(\mK)^{\top}\mV),
% \end{equation}
% where \hat{\boldsymbol{\mu}} \in \mathbb{R}_+^{k}\hat{\boldsymbol{\mu}} \in \mathbb{R}_+^{k} denotes a learnable vector, which will be explained later. 
% \mathrm{ort}(\cdot)\mathrm{ort}(\cdot) signifies an orthogonalization process applied to the features to obtain orthonormal eigenfunctions. 

The non-linear transformation $\sigma$ is implemented following the structure of the traditional attention mechanism, which involves residual connections~\cite{he2016deep} and FFN:
%Residual connections~\cite{he2016deep} and a two-layer FFN transformation is implemented following the orthogonal attention mechanism. The non-linear transformation \sigma\sigma is applied after the first linear layer in the FFN:
% The orthogonal attention layer is defined as:
\begin{equation}
    % \text{Attn}_{self} : \mathbb{R}^{n \times d} \rightarrow \mathbb{R}^{n \times d}, \quad 
    {\bm{h}}^{(l+1)}_{i} = \mathrm{FFN}(\mathrm{LN}(\tilde{\bm{h}}^{(l+1)}_{i} + {\bm{h}}^{(l)}_{i})).
\end{equation}
% Similar to the Galerkin attention encoder layer introduced in \citep{cao2021choose}, our implementation also incorporates layer normalization and a two-layer FFN.
The FFN in the final orthogonal attention serves as $\mathcal{P}$ to map hidden states to PDE solutions.

\textbf{Implementation of  $\mathrm{ort}(\cdot)$.} 
As mentioned, we leverage $\mathrm{ort}(\cdot)$ to make $\hat{\bm{g}}^{(l)}_i$ follow the structure of the outputs of eigenfunctions. 
We highlight that the orthonormalization lies in the function space, i.e., among the output dimensions of the function $\hat{g}^{(l)}: \bm{f}_{i,j} \mapsto \hat{\bm{g}}^{(l)}_{i,j} \in \mathbb{R}^{k}$ instead of the column vectors. 
Thereby, we should not orthonormalize matrix $\hat{\bm{g}}^{(l)}_i$ over its columns but manipulate $\hat{g}^{(l)}$. 

To achieve this, we first estimate the covariance over the output dimensions of $\hat{g}^{(l)}$, which can be approximated by Monte Carlo (MC) estimation:
\begin{equation}
\begin{aligned}
    \mathbf{C}^{(l)} &\approx \frac{1}{NM} \sum_{i=1}^N \sum_{j=1}^M [\hat{g}^{(l)}(\bm{f}_{i,j})^\top \hat{g}^{(l)}(\bm{f}_{i,j})] \\
    &= \frac{1}{NM} \sum_{i=1}^N [\hat{\bm{g}}^{(l)}_i{}^\top \hat{\bm{g}}^{(l)}_i].
\end{aligned}
\end{equation}
Then, we orthonormalize $\hat{g}^{(l)}$ by right multiplying the matrix $\mathbf{L}^{(l)}{}^{-\top}$, where $\mathbf{L}^{(l)}$ is the lower-triangular matrix arising from the Cholesky decomposition of $\mathbf{C}^{(l)}$, i.e., $\mathbf{C}^{(l)} = \mathbf{L}^{(l)} \mathbf{L}^{(l)}{}^\top$. 
In the vector formula, there is
\begin{equation}
    \hat{\bm{\psi}}^{(l)}_i := \hat{\bm{g}}^{(l)}_i \mathbf{L}^{(l)}{}^{-\top}. 
\end{equation}
The covariance of the functions producing $\hat{\bm{\psi}}^{(l)}_i$ can be approximately estimated:
\begin{equation}
% \scriptsize
\begin{aligned}
    \frac{1}{NM} \sum_{i=1}^N &\left[\left(\hat{\bm{g}}^{(l)}_i\mathbf{L}^{(l)}{}^{-\top}\right)^\top \hat{\bm{g}}^{(l)}_i \mathbf{L}^{(l)}{}^{-\top}\right]\\
    &= \mathbf{L}^{(l)}{}^{-1} \mathbf{C}^{(l)} \mathbf{L}^{(l)}{}^{-\top} 
    = \mathbf{I},  
\end{aligned}
\end{equation}
which conforms that these functions can be regarded as orthonormal eigenfunctions that implicitly define a kernel. 

However, in practice, the model parameters evolve repeatedly, we cannot trivially estimate $\mathbf{C}^{(l)}$, which involves the whole training set, at a low cost per training iteration. 
Instead, we propose to approximately estimate $\mathbf{C}^{(l)}$ via 
% We first estimate the covariance among functions defined by ..., 
the exponential moving average trick---similar to the update rule in batch normalization~\cite{ioffe2015batch}, we maintain a buffer tensor $\mathbf{C}^{(l)}$ and update it with training mini-batches. 
We reuse the recorded training statistics to ensure the stability of inference. 
% \begin{equation}
%  \mC = m\mC + (1-m)\mQ^{\top}\mQ
% \end{equation}
% where m∈[0,1]m \in [0,1] is a smoothing factor. 

% However, direct estimation of \mC\mC is not feasible due to the learnable nature of the matrix \mQ\mQ, which evolves during training.

% Instead, we employ  to estimate \mC\mC during training, 
% updating the value with each batch of training samples:

% We implement 
% % To orthogolize \mQ\mQ （\mK\mK, we utilize a covariance matrix \mC\mC and a lower triangular matrix \mL\mL:
% \begin{equation}
%     \mC , \mL \in \mathbb{R}^{k \times k} \quad and \quad \mC := \mQ^{\top}\mQ \quad , \quad \mL = \mathrm{chol}(\mC)
% \end{equation}
% where chol(⋅)\mathrm{chol}(\cdot) represents the Cholesky decomposition yielding a lower triangular matrix. We note that \mL−1\mQ⊤\mQ(\mL−1)⊤=\mIk \mL^{-1}\mQ^{\top}\mQ(\mL^{-1})^{\top} = \mI_{k} . The orthogonalization process can then be performed as follows:
% \begin{equation}
% \mathrm{ort} := \mQ (\mL^{-1})^{\top}.
% \end{equation}

The aforementioned process involves a cubic complexity with respect to $k$ due to the Cholesky decomposition.
However, it is worth noting that empirically, $k$ is significantly smaller than the number of measurement points $M$.
Consequently, the overall complexity of the proposed orthogonal attention mechanism remains moderate (see the empirical results in Table~\ref{tab:runtime}).

%\textbf{Orthogonal attention.} 
%Without loss of generality, we assume that both the input functions and output functions are discretized within the same set of measurement points.
%In the scenario where the discretization of the input functions and solution functions differs, we introduce a cross-attention module. The key change lies in the definition of Q/K/VQ/K/V:
%\begin{equation}
%\mW^{Q} \in \mathbb{R}^{d \times k},\mW^{K} \in \mathbb{R}^{d \times k},\mW^{V} \in \mathbb{R}^{d \times d}  \,  \,  \mQ:=g(\bm{x}') \mW^{Q}, \mK := g(\bm{x}) \mW^{K} , \mV:=\bm{y} \mW^{V}
%\end{equation}
%where g(\bm{x}′)∈Rn′×dg(\bm{x}') \in \mathbb{R}^{n' \times d} and g(\bm{x})∈Rn×dg(\bm{x}) \in \mathbb{R}^{n \times d} represents the learned basis functions discretized within Xout\mathbf{X}_{out} and Xin\mathbf{X}_{in}, respectively, and \mWQ\mW^{Q} is distinct from \mWK\mW^{K}.

%To enhance the expressiveness of our model, we introduce the orthogonal attention layer, elaborated on in section~??????????????????????????????\ref{sec:mode}.
%This layer serves a dual role, functioning as both σl\sigma_{l} and the kernel integral operators Kl\mathcal{K}_{l}.
% \subsection{orthogonal attention mechanism}
% The kernel integral is essential for transforming input functions towards target functions. The theorem below validates the ability of orthogonal attention to approximate the kernel operator.

\subsection{Theoretical Insights} 
\label{sec:theo}
This section provides the theoretical insights behind orthogonal attention. 
We abuse notations when there is no misleading. 
Consider a kernel integral operator $\mathcal{K}$ as follow:
\begin{align}
    (\mathcal{K}h)(\bm{x}) := \int_{D} \kappa(\bm{x}, \bm{x}') h(\bm{x}') \, d\bm{x}', \quad \forall \bm{x} \in D,
\end{align}
where $\kappa$ is a positive semi-definite kernel and $h$ is the input function. 
Given $\psi_i$ as the eigenfunction of $\mathcal{K}$ corresponding to the $i$-th largest eigenvalue $\mu_i$, we have:
\begin{equation}
    \begin{aligned}
    \int_{D} \kappa(\bm{x}, \bm{x}') \psi_{i}(\bm{x}') \, d\bm{x}' &= \mu_i \psi_i(\bm{x}),\; \quad \forall i \geq 1 ,  \forall \bm{x} \in D  \\
    \langle\psi_i, \psi_j\rangle & = \mathbbm{1}[i = j], \quad \forall i, j \geq 1,
    \end{aligned}
\end{equation}
where $\langle a, b\rangle := \int a(\bm{x}) b(\bm{x})\,d\bm{x}$ denotes the inner product in $D$. 
By Mercer's theorem, there is:
% \begin{equation}
%     \kappa(\bm{x}, \bm{x}') = \sum_{i \geq 1} \mu_i \psi_i(\bm{x}) \psi_i(\bm{x}'),
% \end{equation}
\begin{equation}
\label{eq:motivation}
\begin{aligned}
    (\mathcal{K}h)(\bm{x}) &= \int_{D} \kappa(\bm{x}, \bm{x}') h(\bm{x}') \, d\bm{x}' \\
    &= \int \sum_{i \geq 1} \mu_i \psi_i(\bm{x}) \psi_i(\bm{x}') h(\bm{x}') \, d\bm{x}' \\
    &= \sum_{i \geq 1} \mu_i \langle\psi_i, h\rangle \psi_i(\bm{x}).    
\end{aligned}
\end{equation}
% where ⟨ψi,f⟩:=∫ψi(\bm{x}′)f(\bm{x}′)d\bm{x}′\langle\psi_i, f\rangle:=\int \psi_i(\bm{x}') f(\bm{x}') \, d\bm{x}' . 
Although we cannot trivially estimate the eigenfunctions $\psi_i$ in the absence of $\kappa$'s expression, 
Equation \ref{eq:motivation} offers us new insights on how to parameterize a kernel integral operator.
In particular, we can truncate the summation in Equation~\ref{eq:motivation} and introduce a parametric model $\hat{\psi}(\cdot): D \to \mathbb{R}^k$ with orthogonal outputs and build a neural operator $\hat{\mathcal{K}}$ with the following definition:
\begin{equation}
\label{eq:imodel}
    (\hat{\mathcal{K}}h)(\bm{x}) := \sum_{i = 1}^k \langle\hat{\psi}_i, h\rangle \hat{\psi}_i(\bm{x}). 
\end{equation}
We demonstrate the convergence of $\hat{\mathcal{K}}$ towards the ground truth $\mathcal{K}$ under MSE loss in the Appendix~\ref{append:theosup}. 
In practice, we first consider $\mathbf{X} := \{\bm{x}_j\}_{1\leq j \leq M}$ and $\mathbf{Y} := \{\bm{x}_j\}_{1 \leq j \leq M'}$ as two sets of measurement points to discretize the input and output functions. 
% and \mY:={\vyj∈D}1≤j≤Nu\mY := \{ \vy_{j} \in D \}_{1\leq j \leq N_{u} } respectively, 
% and \vui:={ui(\vyj)}1≤j≤Nu\vu_i := \{ u_{i}(\vy_j) \}_{1 \leq j \leq N_{u}}. 
We denote $\hat{\bm{\psi}} \in \mathbb{R}^{M \times k}$ and $\hat{\bm{\psi}}' \in \mathbb{R}^{M' \times k}$ as the evaluation of the model $\hat{\psi}$ on $\mathbf{X}$ and $\mathbf{Y}$ respectively. 
Let $\bm{h} \in \mathbb{R}^{M}$ represent the evaluation of $h$ on $\mathbf{X}$. 
% We use ˆψi(\bm{x})∈RMin×1\hat{\psi}_i(\bm{x}) \in \mathbb{R}^{M_{in} \times 1} and ˆψi(\bm{x}′)∈RMout×1\hat{\psi}_i(\bm{x}') \in \mathbb{R}^{M_{out} \times 1} to denote the ii-th eigenfunction of ˆψ(\bm{x})\hat{\psi}(\bm{x}) and ˆψ(\bm{x}′)\hat{\psi}(\bm{x}') respectively. 
There is: % ⟨ˆψi,hl⟩≈ˆψi(\bm{x})⊤\bm{h}(l)(\bm{x})\langle\hat{\psi}_i, h_{l}\rangle \approx \hat{\psi}_i(\bm{x})^\top \bm{h}^{(l)}(\bm{x}) and :
\begin{equation}
\begin{aligned}
\label{eq:attn}
    (\hat{\mathcal{K}}h)(\mathbf{Y}) \approx \sum_{i = 1}^k [\hat{\psi}_i(\mathbf{X})^\top h(\mathbf{X})] \hat{\psi}_i(\mathbf{Y}) = \hat{\bm{\psi}}' \hat{\bm{\psi}}^\top \bm{h}.
    % &\approx \frac{1}{N} \sum_{n=1}^N \sum_{i = 1}^k \hat{\psi}_i^{\mathbf{X}_{in}}{}^\top f_n^{\mathbf{X}_{in}} \left( \hat{\psi}_i^{\mathbf{X}_{in}}{}^\top f_n^{\mathbf{X}_{in}}  - 2  \hat{\psi}_i^{\mathbf{X}_{out}}{}^\top u_n^{\mathbf{X}_{out}} \right) + C
\end{aligned}
\end{equation}
Comparing Equation~\ref{eq:motivation} and Equation~\ref{eq:imodel}, we can see that the scaling factors $\mu_i$ are omitted, which may undermine the model flexibility in practice. 
To address this, we introduce a learnable vector $\hat{\boldsymbol{\mu}} \in \mathbb{R}_+^{k}$ to Equation~\ref{eq:attn}, resulting in:
\begin{equation}
    (\hat{\mathcal{K}}h)(\mathbf{Y}) \approx \hat{\bm{\psi}}' \mathrm{diag}(\hat{\boldsymbol{\mu}}) \hat{\bm{\psi}}^\top \bm{h}.
\end{equation}
As shown, there is an attention structure---$\hat{\bm{\psi}}' \mathrm{diag}(\hat{\boldsymbol{\mu}}) \hat{\bm{\psi}}^\top$ corresponds to the attention matrix that defines how the output function evaluations attend to the input. 
Besides, the orthonormalization regularization arises from the nature of eigenfunctions, benefitting to alleviate overfitting and boost generalization.
% A comprehensive discussion on this topic can be found in the appendix.
% In this spirit, we provide a sensible explanation for the prevalence of employing attention mechanisms in operator learning.
% With the definitions presented in Equation~\ref{eq:QKV}, we define the orthogonal attention as follows:
% \begin{equation}
%     \label{eq:Attnpro}
%     \mathrm{Attn}_{o} := \mathrm{ort}(\mQ)(\mathrm{ort}(\mK)^{T}\mV).
% \end{equation}
When $\mathbf{X} = \mathbf{Y}$, the above attention structure is similar to regular self-attention mechamism with a symmetric attention matrix.
%, similar to regular self-attention mechanism with a. 
Otherwise, 
% \textbf{The Cross-Attention Variant.}
%\zhijie{in the above paragraph, you do not state that they are discretized with the same points, right? you say Xin and Xout already!}
% For a pair of functions $(\vf_i, \vu_i)$ discretized with different points , 
it boils down to a cross-attention, which enables our approach to query output functions at arbitrary locations independent of the inputs. 
Find more details regarding this in Appendix~\ref{append:theosup}.

\section{Experiments}
We conduct extensive experiments on diverse and challenging benchmarks across various domains to showcase the effectiveness of our method.
%This section compares the proposed method to prior state-of-the-art (SOTA) neural operators on diverse, challenging benchmarks in various domains.

\begin{table*}[t]
\
\caption{The main results on six benchmarks compared with seven baselines. Lower scores indicate superior performance, and the best results are highlighted in bold. ``*" means that the results of the method are reproduced by ourselves. ``-" means that the baseline cannot handle this benchmark.}
\label{table:res}
\begin{center}
\begin{tabularx}{\linewidth}{X >{\centering\arraybackslash}X >{\centering\arraybackslash}X >{\centering\arraybackslash}X >{\centering\arraybackslash}X >{\centering\arraybackslash}X >{\centering\arraybackslash}X}
\toprule
\bf {MODEL}  &\bf{NS2d}  &\bf{Airfoil}  &\bf{Pipe}  &\bf {Darcy}  &\bf{Elasticity}  &\bf{Plasticity}
\\ \midrule 
FNO         &0.1556     &-     &-     &0.0108     &-     &- \\
Galerkin    &0.1401     &-     &-          &0.0084     &-     &- \\
Geo-FNO     &0.1556     &0.0138     &0.0067     &0.0108     &0.0229     &0.0074 \\
U-FNO*       &0.2182     &0.0137     &0.0050     &0.0266     &0.0226     &\textbf{0.0028} \\
F-FNO*       &0.1213     &0.0079     &0.0063     &0.0318     &0.0316     &0.0048 \\
LSM*         &0.1693     &0.0062     &0.0049     &\textbf{0.0069}     &0.0225     &0.0035 \\
GNOT        &0.1380     &0.0076     &-     &0.0105     &\textbf{0.0086}     &- \\
ONO        &\textbf{0.1195}  &\textbf{0.0056}  &\textbf{0.0034}  &0.0072  &0.0118  &0.0048 \\
\bottomrule
\end{tabularx}
\end{center}
\end{table*}

\textbf{Benchmarks.} 
We first evaluate our model's performance on Darcy and NS2d~\cite{li2020fourier} benchmarks to evaluate its capability on regular grids.
Subsequently, we extend our experiments to benchmarks with irregular geometries, including Airfoil, Plasticity, and Pipe, which are represented in structured meshes, as well as Elasticity, presented in point clouds~\cite{li2022fourier}.
%To comprehensively evaluate the effectiveness and adaptability of our approach, we choose a diverse set of benchmarks with distinct geometric representations from both solid and fluid physics.
%For assessing the performance in solving PDEs on regular meshes, we include Darcy and NS2d~\cite{li2020fourier}.
%To gauge the efficiency in handling irregular meshes and point clouds, we incorporate Elasticity, Plasticity, Pipe, and Airfoil as our benchmarks~\cite{li2022fourier}.

\textbf{Baselines.} 
We compare our model with several baseline models, including the well-recognized FNO~\cite{li2020fourier} and its variants Geo-FNO~\cite{li2022fourier}, F-FNO~\cite{tran2023factorized}, and U-FNO~\cite{wen2022u}.
Furthermore, we consider other models such as Galerkin Transformer~\cite{cao2021choose}, LSM~\cite{wu2023LSM}, and GNOT~\cite{hao2023gnot}.
It's worth noting that LSM and GNOT are the latest state-of-the-art (SOTA) neural operators.
%FNO~\cite{li2020fourier}, Geo-FNO~\cite{li2022fourier}, Galerkin Transformer~\cite{cao2021choose}, F-FNO~\cite{tran2021factorized}, U-FNO~\cite{wen2022u}, LSM~\cite{wu2023solving}, and GNOT~\cite{hao2023gnot}.

\textbf{Implementation details.}
We use the $l_{2}$ relative error in Equation~\ref{eq:optim} as the training loss and evaluation metric.
We train all models for 500 epochs.
Our training process employs the AdamW optimizer~\cite{loshchilov2018decoupled} and the OneCycleLr scheduler~\cite{smith2019super}.
We initialize the learning rate at $10^{-3}$ and explore batch sizes within the range of $\{2, 4, 8, 16\}$.
The model's width is set to $128$, while the orthogonalization process employs dimension $d$ as either $8$ or $16$.
Unless specified otherwise, we choose either the Linear Transformer block from~\cite{katharopoulos2020transformers} or the Nystr{\"o}m Transformer block from~\cite{xiong2021nystromformer} as the NN block in our model.
Further implementation details of the baselines are provided in Appendix~\ref{appendix:hyper}.
We also provide a run-time comparison of different neural operators in Table~\ref{tab:runtime}.
Our experiments are conducted on a single NVIDIA RTX 3090 GPU.
%We train all models using the $l_{2}$ loss function for 500 epochs with an initial learning rate of $10^{-3}$.
%We utilize the AdamW optimizer~\cite{loshchilov2017decoupled} and the one-cycle learning rate strategy~\cite{smith2019super}.
%The batch size varies within the range of $\{2, 4, 8, 16, 32\}$.
%Our experiments are conducted on a single NVIDIA RTX 3090 GPU.
%We use the mean $l_{2}$ relative error as the evaluation metric.

\subsection{Main Results}

\begin{figure}[t]
  % \vspace{-5pt}
  \centering
  \includegraphics[width=0.7\linewidth]{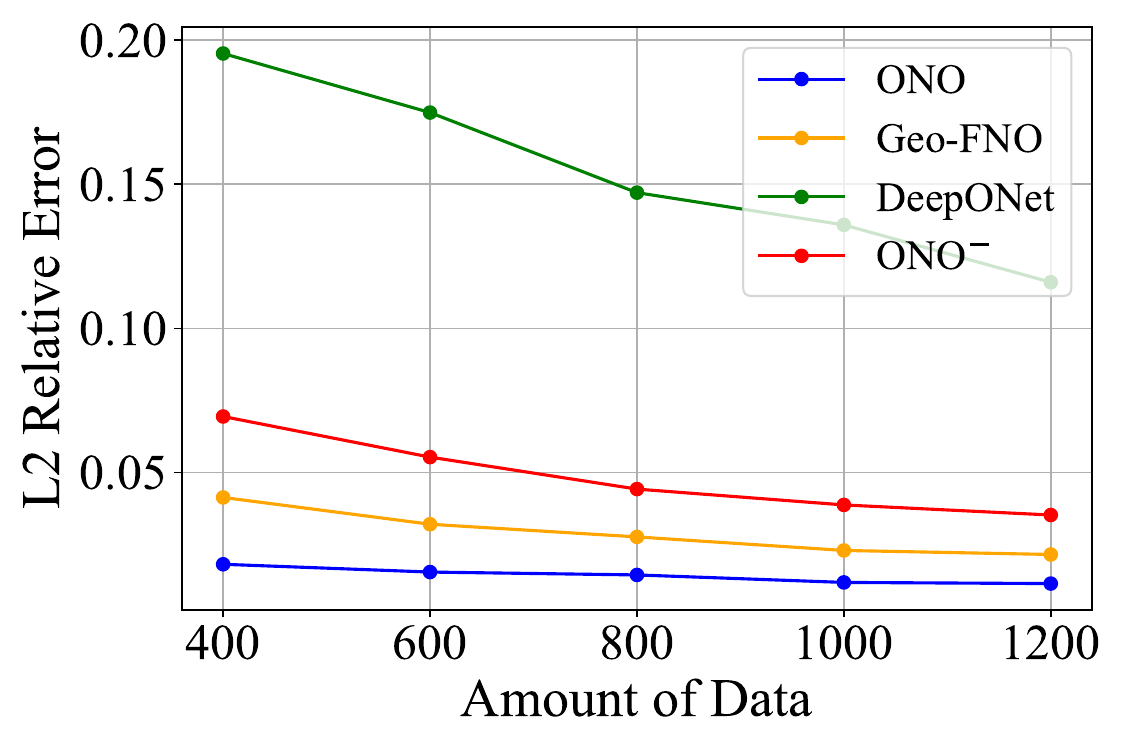}
  \vspace{-10pt}
  \caption{Comparison on the $l_2$ relative error for different training data amounts on Elasticity.}
  \label{fig:dataamount}
  % \vspace{-15pt}
\end{figure}

% We evaluate our model on six diverse datasets, comparing its performance with seven well-established baselines.
%\hzk{specify model parameters in the main Table? If time permits, mark train loss to clarify overfitting in one of the experiments?}
Table~\ref{table:res} reports the results. 
Remarkably, our model achieves SOTA performance on three of these benchmarks, reducing the average prediction error by $13\%$. Specifically, it reduces the error by $31\%$ and $10\%$ on Pipe and Airfoil, respectively.
In the case of NS2d, which involves temporal predictions, our model surpasses all baselines.
We attribute it to the temporal generalization enabled by our orthogonal attention.
We conjecture that the efficacy of orthogonal regularization contributes to our model's excellent performance in these three benchmarks by mitigating overfitting the limited training data.
These three benchmarks encompass both regular and irregular geometries, demonstrating the versatility of our model across various geometric settings.

Our model achieves the second-lowest prediction error on Darcy and Elasticity benchmarks, albeit with a slight margin compared to the SOTA baselines.
We notice that our model and the other attention operator (GNOT) demonstrate a significant reduction in error when compared to other operators that utilize a learnable mapping to convert the irregular geometries into or back from uniform meshes.
This mapping process can potentially introduce errors.
However, attention operators naturally handle irregular meshes for sequence input without requiring mapping, leading to superior performance.
Our model also exhibits competitive performance on plasticity, involving the mapping of a shape vector to the complex mesh grid with a dimension of deformation.
These results highlight the versatility and effectiveness of our model as a framework for operator learning.

\textbf{Training on limited data.} We investigate the influence of limited training data using the Elasticity dataset and make comparisons with FNO and DeepONet, two widely recognized neural operators. 
To demonstrate the effectiveness of the orthogonalization process, we additionally utilize ONO without the orthogonalization, referred to as ONO$^{-}$.

\begin{table}[t]
% \vspace{-10pt}
\centering
\caption{Runtime comparison. ``LT" refers to using the Linear Transformer block for specifying the NN block. 
All models use a batch size of 8. FNO, LSM, and ONO are fixed as 4 layers. 
The width of ONO $d$ is set to 128, and the number of eigenfunctions $k$ is 16.}
\vspace{1ex}
\label{tab:runtime}
\begin{tabular}{lcccc}
\toprule
\bfseries MODEL & \bfseries FNO & \bfseries Galerkin  & \bfseries LSM & \bfseries ONO (LT)  \\
\midrule 
Runtime (s) & 3.81 & 9.88  & 9.05 & 7.87 \\
\bottomrule
\end{tabular}
% \vspace{-15pt}
\end{table}

\begin{table*}
% \hspace{20pt}
\centering
  \begin{minipage}[t]{0.6\linewidth}
    \centering
    \setlength{\tabcolsep}{8pt}
    \caption{Comparison on the $l_2$ relative error for Zero-shot super-resolution on darcy benchmark. s denotes the resolution of the evaluation data. The models are trained on data of $43\times 43$ resolution (s=43).}
    \vspace{1ex}
    \begin{tabularx}{\linewidth}{lccccc}
      \toprule
      {\bf MODEL}  & $s = 61$  & $s = 85$  & $s = 141$  & $s = 211$  & $s = 421$ \\
      \midrule
      FNO     &0.1164   &0.1797   &0.2679   &0.3160   &0.3631 \\
      Ours    &\textbf{0.0204}   &\textbf{0.0259}   &\textbf{0.0315}   &\textbf{0.0349}   &\textbf{0.0386} \\
      \bottomrule
    \end{tabularx}

    \label{table:spatial}
  \end{minipage}%
  \quad\quad% 控制两个表格之间的水平间距
  \begin{minipage}[t]{0.3\linewidth}
    \caption{Comparison on the $l_2$ relative error for seen and unseen timesteps on NS2d. }
    \vspace{1ex}
    \centering
    \setlength{\tabcolsep}{8pt}
    \begin{tabularx}{\linewidth}{lcc}
      \toprule
      {\bf MODEL}  & Seen  &  Unseen \\
      \midrule
      FNO     &0.0982   &0.2446 \\
      Ours    &\textbf{0.0889}   &\textbf{0.2143} \\
      \bottomrule
    \end{tabularx}

    \label{table:temporal}
  \end{minipage}
% \hspace{20pt}
\end{table*}

The results are shown in Figure~\ref{fig:dataamount}.
ONO outperforms the baselines with different training data amounts, followed by Geo-FNO, ONO$^{-}$, and DeepONet.
Each of the neural operators demonstrates a degradation in performance as the training data amount decreases.
The reduction in training data from 1200 to 400 results in significant increases in prediction error for ONO$^{-}$ (97.1$\%$, $0.0352 \rightarrow 0.0694$) and Geo-FNO (92$\%$, $0.0215 \rightarrow 0.0413$). 
DeepONet demonstrates a 68$\%$ increase, while ONO demonstrates the lowest increase of 58$\%$ ($0.0114 \rightarrow 0.0181$).
%DeepONet demonstrates stronger generalization on unseen test data compared to Geo-FNO.
%~\citet{lu2022comprehensive} also reveals that FNOs struggle to effectively handle noisy input.
%These results may stem from FNO's complex parameterization of Fourier modes, leading to overfitting of training data details.
Notably, ONO$^{-}$ exhibits considerable performance deterioration when trained on reduced training data compared to ONO.
The superior generalization ability of ONO, relative to the baselines, highlights the effectiveness of the orthogonalization operation for deep learning-based operator learning. 

\textbf{Runtime comparison}. Table~\ref{tab:runtime} provides a comparison on the runtime of different neural operators, revealing that ONO with a Linear Transformer block has a comparable computational cost to the linear Galerkin Transformer.

\begin{figure}[t]
\centering
  \begin{minipage}[b]{.3\columnwidth}
    \centering
    \includegraphics[width=\linewidth]{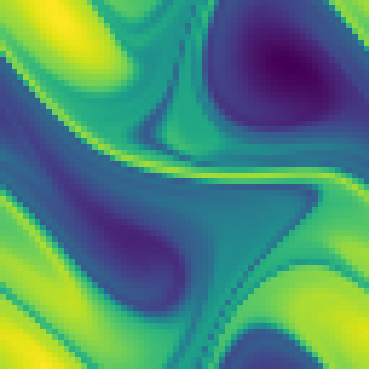}
  \end{minipage}%
  \hspace{1pt}
  \begin{minipage}[b]{.3\columnwidth}
    \centering
    \includegraphics[width=\linewidth]{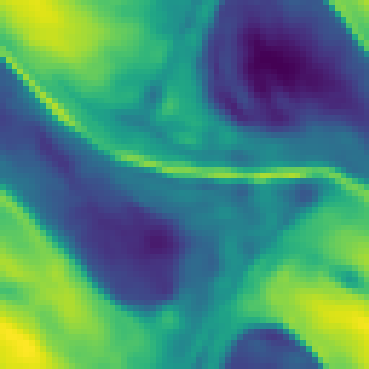}
  \end{minipage}%
  \hspace{1pt}
  \begin{minipage}[b]{.3\columnwidth}
    \centering
    \includegraphics[width=\linewidth]{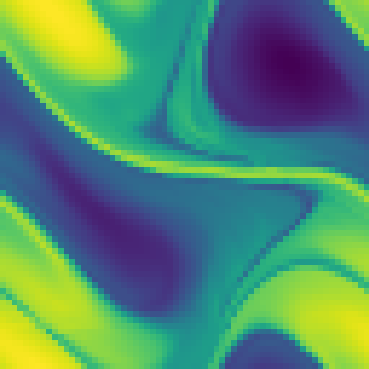}
  \end{minipage}
  \vspace{1pt}  % Add vertical space between rows
  \begin{minipage}[b]{.3\columnwidth}
    \centering
    \includegraphics[width=\linewidth]{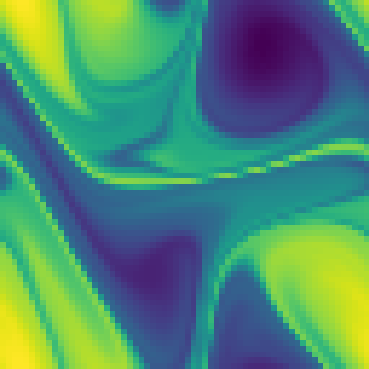}
  \end{minipage}%
  \hspace{1pt}
  \begin{minipage}[b]{.3\columnwidth}
    \centering
    \includegraphics[width=\linewidth]{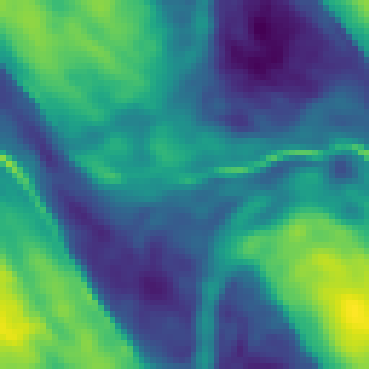}
  \end{minipage}%
  \hspace{1pt}
  \begin{minipage}[b]{.3\columnwidth}
    \centering
    \includegraphics[width=\linewidth]{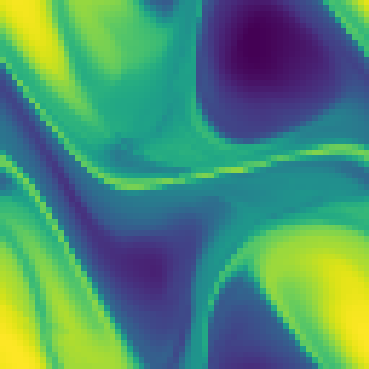}
  \end{minipage}
    
  \caption{The two rows refer to the results of models, trained to predict timesteps 11-18, for timesteps 19 and 20 on NS2d. 
From left to right: ground truth, prediction of FNO, and that of ONO.}
\label{fig:timegen}
\end{figure}

\begin{figure}[t]
  \centering
  %\hspace{55pt}
  \begin{minipage}{0.3\columnwidth}
    \centering
    \includegraphics[width=\linewidth]{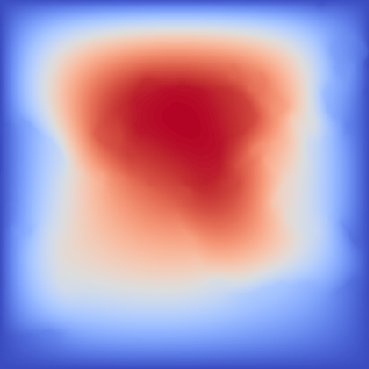}
    \label{fig:subfig1}
  \end{minipage}%
  \hspace{1pt}
  \begin{minipage}{0.3\columnwidth}
    \centering
    \includegraphics[width=\linewidth]{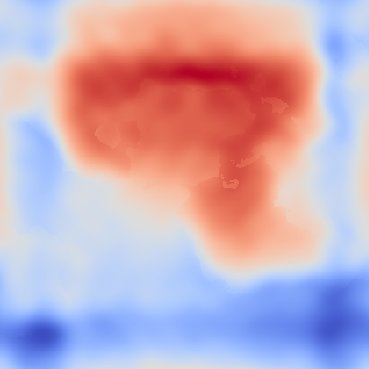}
    \label{fig:subfig2}
  \end{minipage}%
  \hspace{1pt}
  \begin{minipage}{0.3\columnwidth}
    \centering
    \includegraphics[width=\linewidth]{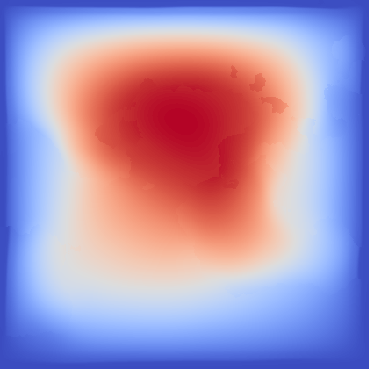}
    \label{fig:subfig3}
  \end{minipage}
  %\hspace{55pt}
  \vspace{-10pt}
  \caption{Zero-shot super-resolution results on Darcy. Models are trained on $43 \times 43$ data and evaluated on $421 \times 421$. From left to right: ground truth, prediction of FNO, and that of ONO. }
  \label{fig:zeroshot}
  \vspace{-5pt}
\end{figure}

\subsection{Generalization Experiments}
We conduct experiments on the generalization performance in both the spatial and temporal axes. 
First, a zero-shot super-resolution experiment is conducted on Darcy.
The model is trained on $43 \times 43$ resolution data and evaluated on resolutions up to nearly ten times that size ($421 \times 421$).
Subsequently, we train the model to predict timesteps 11-18 and evaluate it on two subsequent intervals: timesteps 11-18, denoted as ``Seen", and timesteps 19-20, denoted as ``Unseen".
We choose the FNO~\cite{li2020fourier} as the baseline due to its well-acknowledged mesh-invariant property and is use of the orthogonal Fourier basis functions, which may potentially offer regularization benefits.

The results are shown in Table~\ref{table:spatial} and Table~\ref{table:temporal}. 
On Darcy, the prediction error of FNO increases dramatically as the evaluation resolution grows.
In contrast, our model exhibits a much slower increase in error and maintains a low prediction error even with excessively enlarged resolution,
%Conversely, FNO exhibits an increasing error trend with higher resolutions.
% Our model consistently achieves significantly lower errors across various resolutions, 
notably reducing the prediction error by $89\%$ compared to FNO on the $421 \times 421$ resolution.
%The results showcase the inherent mesh-invariant property of our model.
On NS2d, Our model outperforms in both time intervals, reducing the prediction error by $9 \%$ and $12 \%$.
We further visualize some generalization results in these two scenarios in Figure~\ref{fig:zeroshot} and Figure~\ref{fig:timegen}.
The results are consistent with the reported values. 
These results demonstrate that our model exhibits remarkable generalization capabilities in both temporal and spatial domains. 
Acquiring high-resolution training data can be computationally expensive.
Our model's mesh-invariant property enables effective high-resolution performance after being trained on low-resolution data, potentially resulting in significant computational cost savings.

\begin{figure*}
    \centering
    %\vspace{-0ex}
    \begin{minipage}{0.35\linewidth}
        \centering
        \includegraphics[width=\linewidth]{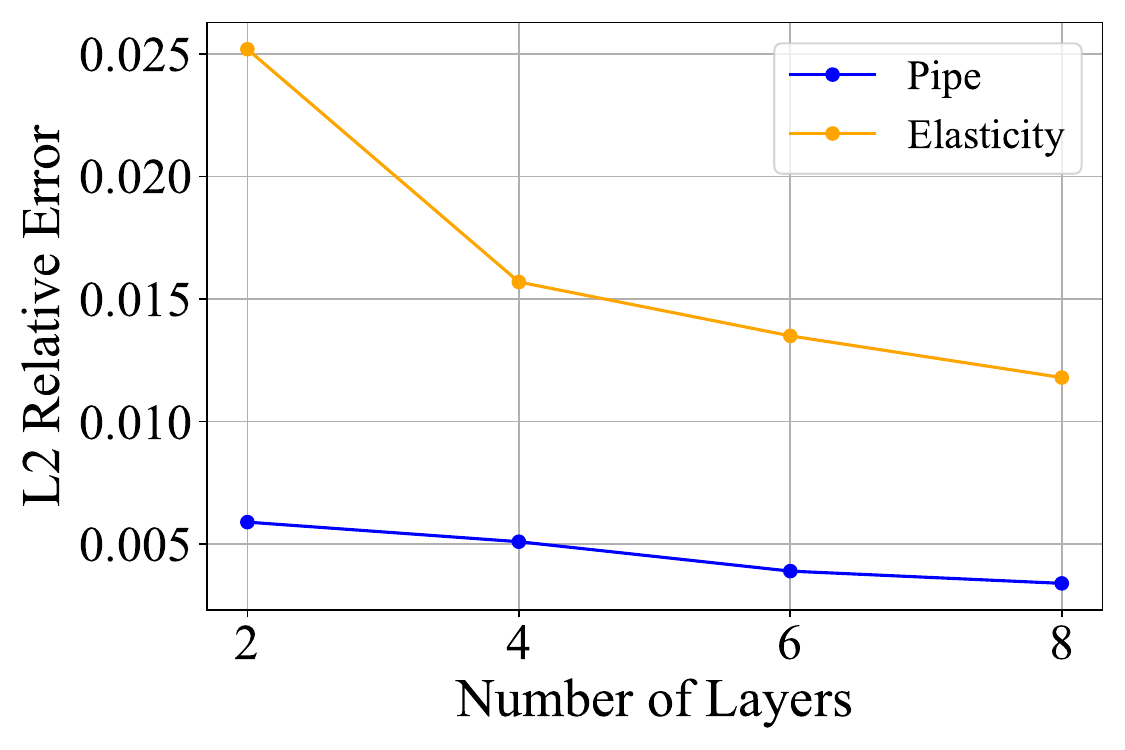}
        %\caption{}
    \end{minipage}
    %\hfill
    \hspace{35pt}
    \begin{minipage}{0.35\linewidth}
        \vspace{10pt}
        \centering
        \includegraphics[width=\linewidth]{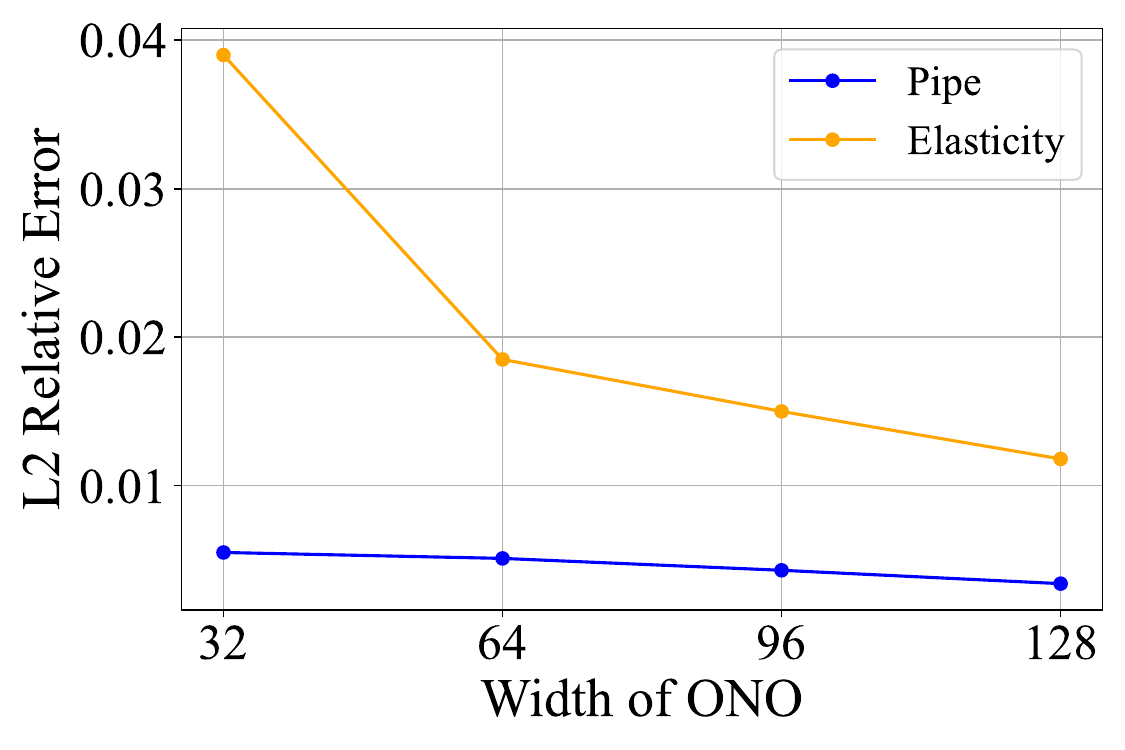}
        %\caption{}
        \label{fig:width}
    \end{minipage}
    %\hspace{0.01\linewidth}
    \vspace{-15pt}
    \caption{$l_2$ relative error varies w.r.t. the number of layers (Left) and width (right) of ONO on Pipe and Elasticity benchmarks. }
    \label{fig:abla}
    \vspace{-10pt}
\end{figure*}

\subsection{Ablation experiments}
To assess the effectiveness of various components in ONO, we conduct a comprehensive ablation study on three benchmarks: Airfoil, Elasticity, and Pipe.

\textbf{Influence of NN Block.}
To show the compatibility of our model, we conduct experiments with different NN blocks.
We choose the Galerkin Transformer block in operator learning~\cite{cao2021choose} and two linear transformer blocks in other domains, including the Linear Transformer block in~\cite{xiong2021nystromformer} and the Nystr{\"o}m Transformer block in~\cite{katharopoulos2020transformers}. 

\begin{table}[t]
\centering
\vspace{-5pt}
\setlength{\tabcolsep}{4pt}
\caption{Comparison on the $l_2$ relative error for different NN blocks on Airfoil, Elasticity, and Pipe benchmarks.}  
\vspace{1ex}
\label{table:nnblock}
\setlength{\tabcolsep}{10pt}
%\vspace{ex}
\begin{tabular}{lccc}
  \toprule
  {\bf DESIGNS}  & Airfoil  & Elasticity & Pipe \\
  \midrule
  Linear     &0.0079   &0.0137   &0.0060 \\
  Nystrom    &\textbf{0.0056}   &\textbf{0.0118}   &\textbf{0.0034} \\
  Galerkin   &0.0122   &0.0133   &0.0075 \\
  \bottomrule
\end{tabular}
\vspace{-5pt}
\end{table}

Table~\ref{table:nnblock} showcases the results. The Nystr{\"o}m Transformer block performs better on all three benchmarks and reduces the error up to 43$\%$ on Pipe.
Linear Transformer block exhibits superior performance on Airfoil and Pipe compared to the Galerkin Transformer block while demonstrating similar performance on Elasticity.
We notice that the Linear Transformer block and Galerkin Transformer block are both kernel-based methods transformer methods.
The Nystr{\"o}m attention uses a downsampling approach to approximate the softmax attention, which aids in capturing positional relationships and contributes to the feature extraction.
However, all the variants consistently exhibit competitive performance, showcasing the flexibility and robustness of our model.

\begin{table}[t] 
\centering
% \vspace{-10pt}
\caption{Comparison on the $l_2$ relative error for orthogonalization and normalization techniques on Airfoil, Elasticity, and Pipe.}
\vspace{1ex}
\label{table:ort}
\setlength{\tabcolsep}{10pt}
\begin{tabular}{lccc}
  \toprule
  {\bf DESIGNS}  & Airfoil  & Elasticity  & Pipe \\
  \midrule
  BN       &0.0808   &0.0149   &0.2151   \\
  LN       &0.0288   &0.0387   &0.0056   \\
  Ortho    & \textbf{0.0056}   & \textbf{0.0118}   & \textbf{0.0034}  \\
  \bottomrule
\end{tabular}
\vspace{-10pt}
\end{table}%

\textbf{Influence of Orthogonalization.}
%A significant distinguishing characteristic of our orthogonal attention mechanism, as opposed to other attention mechanisms, is the inclusion of the orthogonalization process.
To further investigate the impact of the orthogonalization process, we carry out a series of experiments on three benchmarks.
``BN" and ``LN" denote the batch normalization~\cite{ioffe2015batch} and the layer normalization~\cite{ba2016layer}, while ``Ortho" signifies the orthogonalization process in the attention module.
It's worth noting that the attention mechanism coupled with layer normalization assumes a structure resembling Fourier-type attention~\cite{cao2021choose}.

As shown in Table~\ref{table:ort}, our orthogonal attention consistently outperforms other attention mechanisms across all benchmarks, resulting in a remarkable reduction of prediction error, up to $81\%$ on Airfoil and $39\%$ on Pipe.
We conjecture that the orthogonalization may benefit model training through feature scaling.
Additionally, the inherent linear independence among orthogonal eigenfunctions aids the model in effectively distinguishing between various features, contributing to its superior performance compared to conventional normalizations.

\subsection{Scaling Experiments}
Our model's architecture offers scalability, allowing adjustments to both its depth and width for enhanced performance or reduced computational costs.
We conduct scaling experiments to examine how the prediction error changes with the number of layers and the width.

Figure~\ref{fig:abla} shows the results.
The left one depicts the change in prediction error with an increasing number of layers, while the right one shows how the error responds to a growth in the width of ONO.
It is evident that error reduction correlates positively with both the number of layers and width.
Nevertheless, diminishing returns become apparent when exceeding four layers or a width of 64 on Elasticity.
We recommend employing a model configuration consisting of four layers and a width of 64 due to its favorable balance between performance and computational cost.

\begin{table}[t] % 指定表格的位置（r：右侧），宽度为文本宽度的30%
    \centering
    % \small
    % \vspace{10pt}
    \setlength{\tabcolsep}{10pt}
    \caption{Comparison on the $l_2$ relative error for ONO with different depths on Elasticity and Plasticity benchmarks.}
    \vspace{1ex}
    \begin{tabular}{ccc}
    \toprule
    \bf{Model} & Elasticity & Plasticity \\
    \midrule
    ONO-8  & 0.0118 & 0.0048 \\
    ONO-30 & \bf{0.0047} & \bf{0.0013} \\
    \bottomrule
    \end{tabular}
    \label{table:ono30}
    \vspace{-10pt}
\end{table}

To further assess the scalability of our model, we increase the number of layers to 30 and the learnable parameters to 10 million while keeping the width at 128. 
We compare it to the 8-layer model, which has approximately 1 million parameters.
The results are in Table~\ref{table:ono30}.
We denote the models as ``ONO-30" and ``ONO-8" respectively.
The prediction of ONO-30 exhibits a remarkable decrease in both benchmarks, achieving reductions of $37\%$ and $76\%$.
The results demonstrate the potential of ONO as a large pre-trained neural operator.
%We assess performance using limited training data in the context of the Elasticity benchmark, as depicted in Figure~\ref{fig:data}.
%We choose Geo-FNO~\cite{li2022fourier} and DeepONet~\cite{lu2019deeponet}, two well-studied neural operators, as our baselines.
%ONO consistently outperforms the baselines and exhibits a smaller decrease in performance as the amount of training data decreases.
%Remarkably, even with only $25\%$ of the data used, ONO achieves a similar relative error (0.0181 with 400 training data) compared to Geo-FNO (0.0184 with 1600 training data).
%These results highlight the inherent regularization capability of ONO, which effectively prevents overfitting on limited data.

\section{Conclusion}
This paper aims to address the performance decline stemming from the limited training data from classical PDE solvers and the complexity of deep models.
Our main contribution is the introduction of regularization mechanisms for neural operators, which effectively enhance generalization performance with reduced training data.
We propose an attention mechanism with orthogonalization regularization based on the kernel integral rewritten by orthonormal eigenfunctions.
We further present a neural operator called ONO, built upon this attention mechanism.
Extensive experiments demonstrate the superiority of our approach compared to baselines.
The study aims to mitigate the challenges associated with the small data regime and enhance the robustness of large PDE-solving models.
\section*{Acknowledgements}
This work was supported by NSF of China (No. 62306176), Natural Science Foundation of Shanghai (No. 23ZR1428700), Key R\&D Program of Shandong Province, China (2023CXGC010112), and CCF-Baichuan-Ebtech Foundation Model Fund.

\section*{Impact Statement}
This work introduces a neural operator designed to effectively solve PDEs, which is of significance in scientific and engineering domains.
As a foundational machine learning research, the immediate negative consequences are not evident, and the risk of misuse is currently low.

\nocite{langley00}

\bibliography{example_paper}
\bibliographystyle{icml2024}

%%%%%%%%%%%%%%%%%%%%%%%%%%%%%%%%%%%%%%%%%%%%%%%%%%%%%%%%%%%%%%%%%%%%%%%%%%%%%%%
%%%%%%%%%%%%%%%%%%%%%%%%%%%%%%%%%%%%%%%%%%%%%%%%%%%%%%%%%%%%%%%%%%%%%%%%%%%%%%%
% APPENDIX
%%%%%%%%%%%%%%%%%%%%%%%%%%%%%%%%%%%%%%%%%%%%%%%%%%%%%%%%%%%%%%%%%%%%%%%%%%%%%%%
%%%%%%%%%%%%%%%%%%%%%%%%%%%%%%%%%%%%%%%%%%%%%%%%%%%%%%%%%%%%%%%%%%%%%%%%%%%%%%%
\newpage
\appendix
\onecolumn

\section{Theoretical supplement}
\label{append:theosup}
\textbf{Proof of the convergence of $\hat{\mathcal{K}}$.}
To push $\hat{\mathcal{K}}$ in Equation~\ref{eq:imodel} towards to unknown ground truth $\mathcal{K}$, we solve the following minimization problem:
\begin{equation}
\begin{aligned}
    \min_{\hat{\psi}} \ell, \; l&:= \mathbb{E}_{h\sim p(h)} \left( \int \Big[\sum_{i = 1}^k \langle\hat{\psi}_i, h\rangle\hat{\psi}_i(\bm{x})  - (\mathcal{K} h)(\bm{x})\Big]^2 \, d \bm{x} \right)\\
    s.t.: &\,\langle \hat{\psi}_i, \hat{\psi}_j \rangle = \mathbbm{1}[i=j], \quad \forall i, j \in [1, k],
\end{aligned}
\end{equation}

We next prove that the above learning objective closely connects to eigenfunction recovery. 
To show that, we first reformulate the above loss:
\begin{equation}
\begin{aligned}
    \ell &= \mathbb{E}_{h\sim p(h)} \left( \sum_{i = 1}^k \sum_{i' = 1}^k \langle\hat{\psi}_i, h\rangle \langle\hat{\psi}_{i'}, h\rangle \langle \hat{\psi}_i, \hat{\psi}_{i'}\rangle - 2 \sum_{i = 1}^k \langle\hat{\psi}_i, h\rangle \langle\hat{\psi}_i, \mathcal{K} h\rangle + C \right)\\
    &= \mathbb{E}_{h\sim p(h)} \left( \sum_{i = 1}^k \sum_{i' = 1}^k \langle\hat{\psi}_i, h\rangle \langle\hat{\psi}_{i'}, h\rangle \mathbbm{1}[i=i'] - 2 \sum_{i = 1}^k \langle\hat{\psi}_i, h\rangle \langle\hat{\psi}_i, \mathcal{K} h\rangle + C \right)\\
    &= \mathbb{E}_{h\sim p(h)} \left( \sum_{i = 1}^k \langle\hat{\psi}_i, h\rangle^2  - 2 \sum_{i = 1}^k \langle\hat{\psi}_i, h\rangle \langle\hat{\psi}_i, \mathcal{K} h\rangle + C \right)\\
    &= \mathbb{E}_{h\sim p(h)} \left( \sum_{i = 1}^k  \langle\hat{\psi}_i, h\rangle^2  - 2 \sum_{i = 1}^k \langle\hat{\psi}_i, h\rangle \left[\sum_{j \geq 1} \mu_j \langle\psi_j, h\rangle \langle\hat{\psi}_i, \psi_j\rangle\right]  + C\right) \\
\end{aligned}
\end{equation}
where $C$ denotes a constant agnostic to $\hat{\psi}$.

Represent $\hat{\psi}_i$ with its coordinates $\bm{a}_i :=[\bm{a}_{i,1},\dots, \bm{a}_{i,j},\dots]$ in the space spanned by $\{\psi_j\}_{j\geq 1}$, i.e., $\hat{\psi}_i = \sum_{j \geq 1} \bm{a}_{i, j} \psi_j$. 
Thereby, $\langle \hat{\psi}_i, \hat{\psi}_{i'} \rangle = \bm{a}_i^\top \bm{a}_{i'} := \sum_{j\geq 1} \bm{a}_{i, j} \bm{a}_{i', j}$ and $\bm{a}_i^\top \bm{a}_{i'} = \mathbbm{1}[i = i']$. 
Likewise, we represent $h$ with coordinates $\bm{a}_h :=[\bm{a}_{h,1},\dots, \bm{a}_{h,j},\dots]$. 
Let $\boldsymbol{\mu} := [\bm{u}_1, \bm{u}_2, \dots]$ and $\bm{a}_h := \mathbb{E}_{h\sim p(h)}\bm{a}_h \bm{a}_h^\top$. 
There is (we omit the above constant)
\begin{equation}
    \begin{aligned}
        \ell &= \mathbb{E}_{h\sim p(h)} \left( \sum_{i = 1}^k  (\bm{a}_i^\top \bm{a}_h)^2  - 2 \sum_{i = 1}^k (\bm{a}_i^\top \mathrm{a}_h) \left[\sum_{h \geq 1} \mu_j \bm{a}_{h,j} \bm{a}_{i, h}\right]  \right) \\
        &= \mathbb{E}_{h\sim p(h)} \left( \sum_{i = 1}^k  (\bm{a}_i^\top \bm{a}_h)^2  - 2 \sum_{i = 1}^k (\bm{a}_i^\top \bm{a}_h)  (\bm{a}_{i}^\top  \mathrm{diag}(\boldsymbol{\mu}) \bm{a}_{h})  \right) \\ 
        &= \mathbb{E}_{h\sim p(h)} \left( \sum_{i = 1}^k  \bm{a}_i^\top (\bm{a}_h \bm{a}_h^\top) \bm{a}_i  - 2 \sum_{i = 1}^k \bm{a}_i^\top (\bm{a}_h \bm{a}_h^\top) \mathrm{diag}(\boldsymbol{\mu}) \bm{a}_i  \right) \\ 
        &=  \sum_{i = 1}^k \bm{a}_i^\top \left[\mathbb{E}_{h\sim p(h)}\bm{a}_h \bm{a}_h^\top\right] \bm{a}_i - 2 \sum_{i = 1}^k \bm{a}_i^\top \left[\mathbb{E}_{h\sim p(h)}\bm{a}_h \bm{a}_h^\top\right] \mathrm{diag}(\boldsymbol{\mu}) \bm{a}_i \\ 
        &=  \sum_{i = 1}^k  \left[\bm{a}_i^\top \bm{a}_h \bm{a}_i - 2  \bm{a}_i^\top \bm{a}_h \mathrm{diag}(\boldsymbol{\mu}) \bm{a}_i\right] \\
        &=  \sum_{i = 1}^k  \left[\bm{a}_i^\top \bm{a}_h \bm{a}_i - \bm{a}_i^\top \bm{a}_h \mathrm{diag}(\boldsymbol{\mu}) \bm{a}_i - \bm{a}_i^\top  \mathrm{diag}(\boldsymbol{\mu})\bm{a}_h \bm{a}_i \right] \\
        &=  \sum_{i = 1}^k \bm{a}_i^\top \left[\bm{a}_h  - \bm{a}_h \mathrm{diag}(\boldsymbol{\mu}) - \mathrm{diag}(\boldsymbol{\mu})\bm{a}_h\right] \bm{a}_i. 
    \end{aligned}
\end{equation}
$\bm{a}_h$ and $\bm{a}_h  - \bm{a}_h \mathrm{diag}(\boldsymbol{\mu}) - \mathrm{diag}(\boldsymbol{\mu})\bm{a}_h$ are both symmetric positive semidefinite matrices with infinity rows and columns. 
Considering the orthonoramlity constraint on $\{\bm{a}_i\}_{i=1}^k$, minimizing $\ell$ will push $\{\bm{a}_i\}_{i=1}^k$ towards the $k$ eigenvectors with smallest eigenvalues of $\bm{a}_h -  \bm{a}_h \mathrm{diag}(\boldsymbol{\mu}) - \mathrm{diag}(\boldsymbol{\mu}) \bm{a}_h$. 
In the case that $\bm{a}_h$ equals to the identity matrix, i.e., $\mathbb{E}_{h\sim p(h)} \langle h, \psi_i \rangle \langle h, \psi_j \rangle = \mathbbm{1}[i=j]$, there is :
\begin{equation}
\ell = \sum_{i = 1}^k \bm{a}_i^\top \left[\bm{a}_h  - \bm{a}_h \mathrm{diag}(\boldsymbol{\mu}) - \mathrm{diag}(\boldsymbol{\mu})\bm{a}_h\right] \bm{a}_i = k - 2 \sum_{i=1}^k \bm{a}_i^\top \mathrm{diag}(\boldsymbol{\mu}) \bm{a}_i. 
\end{equation}
Then $\{\bm{a}_i\}_{i=1}^k$ will converge to the $k$ principal eigenvectors of $\mathrm{diag}(\boldsymbol{\mu})$, i.e., the one-hot vectors with $i$-th element equaling $1$. 
Given that $\hat{\psi}_i = \sum_{j \geq 1} \bm{a}_{i, j} \psi_j$, the deployed parametric model $\hat{\psi}$ will converge to the $k$ principal eigenfunctions of the unknown ground-truth kernel integral operator.

\textbf{Cross-attention Variant.}
For a pair of functions $(\bm{f}_n, \bm{u}_n)$, the data points used to discretize them are different, denoted as $\mathbf{X}$ and $\mathbf{Y}$.
Let $\mathcal{H}^{(l)}, l\in[1,L]$ denote the specified operators at various modeling stages.
We define the propagation rule as
\begin{equation}
    \begin{aligned}
    (\mathcal{H}^{(1)}\bm{h}^{(1)})(\mathbf{Y}) &\approx \mathrm{FFN}(\mathrm{LN}(\left(\hat{\bm{\psi}}^{(1)}(\mathbf{Y}) \hat{\bm{\psi}}^{(1)}(\mathbf{X})^\top \left[\bm{h}^{(1)}(\mathbf{X})\right] \right)))\\
    (\mathcal{H}^{(l)}\bm{h}^{(l)})(\mathbf{Y}) &\approx \mathrm{FFN}(\mathrm{LN}(\left(\hat{\bm{\psi}}^{(l)}(\mathbf{Y}) \hat{\bm{\psi}}^{(l)}(\mathbf{Y})^\top \left[\bm{h}^{(l)}(\mathbf{Y})\right] + \bm{h}^{(l)}(\mathbf{Y}) \right))),\; l\in[2,L] 
    \end{aligned}
\end{equation}
where $\mathrm{FFN(\cdot)}$ denotes a two-layer FFN and $\mathrm{LN}(\cdot)$ denotes the layer normalization.
\section{Hyperparameters and Details for Models}
\label{appendix:hyper}
\textbf{FNO and its Variants.}
For FNO and its variants (Geo-FNO, F-FNO, U-FNO), we employ 4 layers with modes of $12$ and widths from $\{20,32\}$.
Notably, Geo-FNO reverts to the vanilla FNO when applied to benchmarks with regular grids, resulting in equivalent performance for Darcy and NS2d benchmarks.
For U-FNO, the U-Net path is appended in the last layer.
FNO-2D is implemented in generalization experiments.
The batch size is selected from $\{10,20\}$.

\textbf{LSM.}
We configure the model with 8 basis operators and 4 latent tokens. The width of the first scale is set to 32, and the downsampling ratio is 0.5. The batch size is selected from $\{10, 20\}$.

\begin{table}[h]
\centering
\caption{Comparison of parameter count and memory requirements between ONO and baseline models.}
\label{tab:appcom}
\begin{tabular}{lcccccc}
\toprule
{\bf MODEL} & FNO & U-FNO & LSM & Galerkin & ONO (Linear) & ONO (Nystr{\"o}m)\\
\midrule
Params (M)& 0.9-18.9 & 1.0-19.4 & 4.8-13.9 & 2.2-2.5 & 0.8-2.0 & 0.8-2.0 \\
Memory (G)& 1-3 & 1-4 & 1-6 & 2-9 & 2-14 & 	8-16 \\
\bottomrule
\end{tabular}
\end{table}

\section{Limitation.}
One limitation of this study is the mamory requirement as shown in Table~\ref{tab:appcom}.
As shown, despite with comparable or fewer parameters, ONO exhibits higher memory requirements than the baselines, which is attributed to its dual-pathway architecture. 
To mitigate the memory overhead, we can properly lighten the lower pathway of ONO.
We can also include sub- and up-sampling mechanisms in the front and end of ONO to shorten the sequence length for memory reduction.
%%%%%%%%%%%%%%%%%%%%%%%%%%%%%%%%%%%%%%%%%%%%%%%%%%%%%%%%%%%%%%%%%%%%%%%%%%%%%%%
%%%%%%%%%%%%%%%%%%%%%%%%%%%%%%%%%%%%%%%%%%%%%%%%%%%%%%%%%%%%%%%%%%%%%%%%%%%%%%%

\end{document}